\documentclass[a4paper,fleqn]{cas-dc}
\usepackage[sort&compress,numbers]{natbib}
\usepackage{framed,multirow}
\usepackage{url}
\usepackage{xcolor}
\usepackage{pifont}
\usepackage{graphicx}
\usepackage{bbm}
\usepackage{diagbox}
\usepackage{colortbl}

\newcommand{\cmark}{\checkmark} % Checkmark
\newcommand{\xmark}{\text{\sffamily\large\ding{55}}} % Cross mark

\def\tsc#1{\csdef{#1}{\textsc{\lowercase{#1}}\xspace}}
\tsc{WGM}
\tsc{QE}

\begin{document}
\let\WriteBookmarks\relax
\def\floatpagepagefraction{1}
\def\textpagefraction{.001}

\shorttitle{Causal-Ex}    

\shortauthors{Tan et al.}  
\fntext[fn1]{Preprint uploaded to arXiv. The paper is under consideration at Pattern Recognition Letters.}

\title[mode = title]{\textsf{Causal-Ex}: Causal Graph-based Micro and Macro Expression Spotting}

\author{Pei-Sze Tan*} 
\cortext[cor1]{Corresponding author}
\ead{tan.peisze@monash.edu}

\author{Sailaja Rajanala} 
\ead{sailaja.rajanala@monash.edu}

\author{Arghya Pal}
\ead{arghya.pal@monash.edu}

\author{Rapha\"{e}l C.-W. Phan}
\ead{raphael.phan@monash.edu}

\author{Huey-Fang Ong}
\ead{ong.hueyfang@monash.edu}

\affiliation{organization={CyPhi AI Lab, Monash University, Malaysia campus},
country={Malaysia}}

\begin{abstract}
   Detecting concealed emotions within apparently normal expressions is crucial for identifying potential mental health issues and facilitating timely support and intervention. The task of spotting macro and micro-expressions involves predicting the emotional timeline within a video, accomplished by identifying the onset, apex, and offset frames of the displayed emotions. Utilizing foundational facial muscle movement cues, known as facial action units, boosts the accuracy. However, an overlooked challenge from previous research lies in the inadvertent integration of biases into the training model. These biases arising from datasets can spuriously link certain action unit movements to particular emotion classes. We tackle this issue by novel replacement of action unit adjacency information with the action unit causal graphs. This approach aims to identify and eliminate undesired spurious connections, retaining only unbiased information for classification. Our model, named \textsf{Causal-Ex} (\textbf{Causal}-based \textbf{Ex}pression spotting), employs a rapid causal inference algorithm to construct a causal graph of facial action units. This enables us to select causally relevant facial action units. Our work demonstrates improvement in overall F1-scores compared to state-of-the-art approaches with 0.388 on CAS(ME)$^2$ and 0.3701 on SAMM-Long Video datasets.
\end{abstract}


\maketitle

\section{Introduction}
Facial expression spotting and recognition are intricately woven tasks. However, spotting forms the foundation of any recognition system~\cite{li2017towards}. Expression spotting entails the challenging task of identifying the frames that are most likely to convey emotions. These expressions can be very obvious, otherwise dubbed as macro expressions, or subtly hidden, namely micro-expressions.

The occurrence of micro-expressions is momentary, and the corresponding intensity of facial muscle movements is low. Consequently, it is highly challenging to identify the onset, offset, and apex frames at which micro-expressions reach their peak~\cite{yu2022facial}. Tasks involving the spotting of both macro and micro-expressions have applications in psychology, human-computer interaction, emotion recognition, lie detection, and various other domains. 
\begin{figure}
    \centering
    \includegraphics[width=0.47\textwidth]{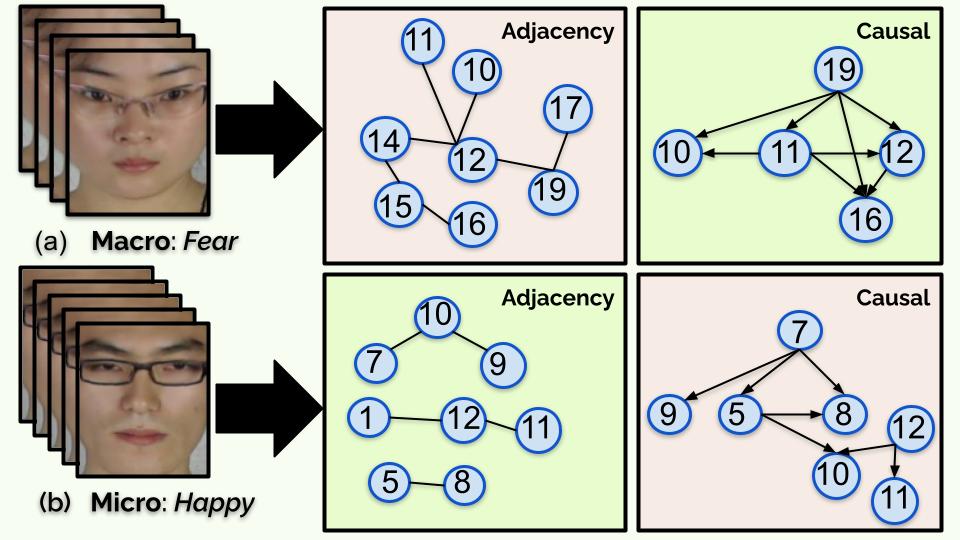}
    \caption{The image displayed above illustrates two sample videos labeled as (a) and (b). Accompanying them are their respective graphs depicting Region of Interest (ROI) adjacency and ROI causal relations. These causal graphs effectively represent the causal links between different ROIs. They play a crucial role in analyzing and comprehending the ways specific ROIs impact others, ultimately contributing to the visualization of emotional expressions. On the other hand, the adjacency graphs primarily showcase patterns of ROI co-occurrence, albeit with less informative value. These patterns might encompass considerable spurious or duplicated information.  }
    \label{fig:intro}
    \vspace{-0.5cm}
\end{figure}
Previous literature has demonstrated the use of long short-term memory (LSTM) ~\cite{lstmspot} to capture the temporal correlations within sequences of videos and optical flow features to detect the rate of change of these features~\cite{liong2021shallow,he2022micro,zhao2022rethinking}. Although the emerging graph-based methods~\cite{yin2023auaware} aims to uncover the intrinsic semantic connections within facial action units, they overlook the potential causal relationships that could establish cause-and-effect associations between groups of action units and resulting emotions. Figure~\ref{fig:intro} illustrates an instance showcasing the contrast between the action unit~(AU) co-occurrence-based adjacency matrix and the AU causal graph. The adjacency-based approaches solely emphasize co-occurrence patterns, which could arise by chance, without accounting for the cause-and-effect associations among the action units. Moreover, the conventional techniques can be prone to noise, struggle with capturing delicate motions, and disregard the inadvertent inclusion of biases linked to data~\cite{li2020deeper}.
The dataset biases have been known to be associated with imbalanced distributions of the training data w.r.t an emotion class, gender, age or any other latent factors~\cite{li2020deeper}. 

In this study, we address two challenges. Firstly, we use structural causal graphs to identify the causal semantics among facial regions' feature space. Secondly, we devise an effective approach to assess the model bias using strategic counterfactuals, which effectively reveal the 
hidden biases.  The proposed counterfactual debiasing process is conducted post-training and is notably faster and simpler to execute.

To overcome the challenges and limitations in macro and micro-expression spotting of the previous works, we introduce a model named \textsf{Causal-Ex}, a Bias-aware AU \underline{Causal} Graph-based facial \underline{Ex}pression detector. This model possesses the capability to unveil causal relationships among AUs using minimal visual data, while also having the potential to mitigate dataset bias. 
 Our contributions in this regard are three-fold:

\begin{itemize}
    \item \underline{\textbf{Causal AU Relation Graph}}: We present an approach to generate AU-based causal relation graphs from a provided video, employing a causal structure discovery algorithm~\cite{fci}. This stands in contrast to existing methods that depend on experts and pre-existing information to manually construct the action unit relation graph~\cite{yin2023auaware}.
    
    \item \underline{\textbf{Improved Spatial-Temporal Embedding Architecture}}: We introduce an enhanced framework for macro and micro-expression spotting, featuring a spatiotemporal graph embedding module designed to capture the complexities of emotional transitions. To improve the accumulation of temporal knowledge, we incorporate a graph convolutional network (GCN). The model's ability to grasp subtle micro-expressions is further supported by maxpooling layers within the convolutional network.

    \item \underline{\textbf{Counterfactual Debiasing}}: We frame the process of debiasing as counterfactual inference~\cite{qian2021counterfactual}. Initially, counterfactuals are employed to interrogate the model, exposing its inherent biased decisions. Subsequently, a debiasing procedure is applied conditionally to rectify the model's learned biases.
\end{itemize}
After ablation analysis, our experiments show that the proposed model learns effectively with the causal graph structures that are computed from the input videos. The results outperform the baselines and the robustness of the model is demonstrated by testing with cross-database generalization capability.  

\begin{figure*}
   \centering
    \hspace{-0.5cm}
    \includegraphics[width=.8\textwidth]{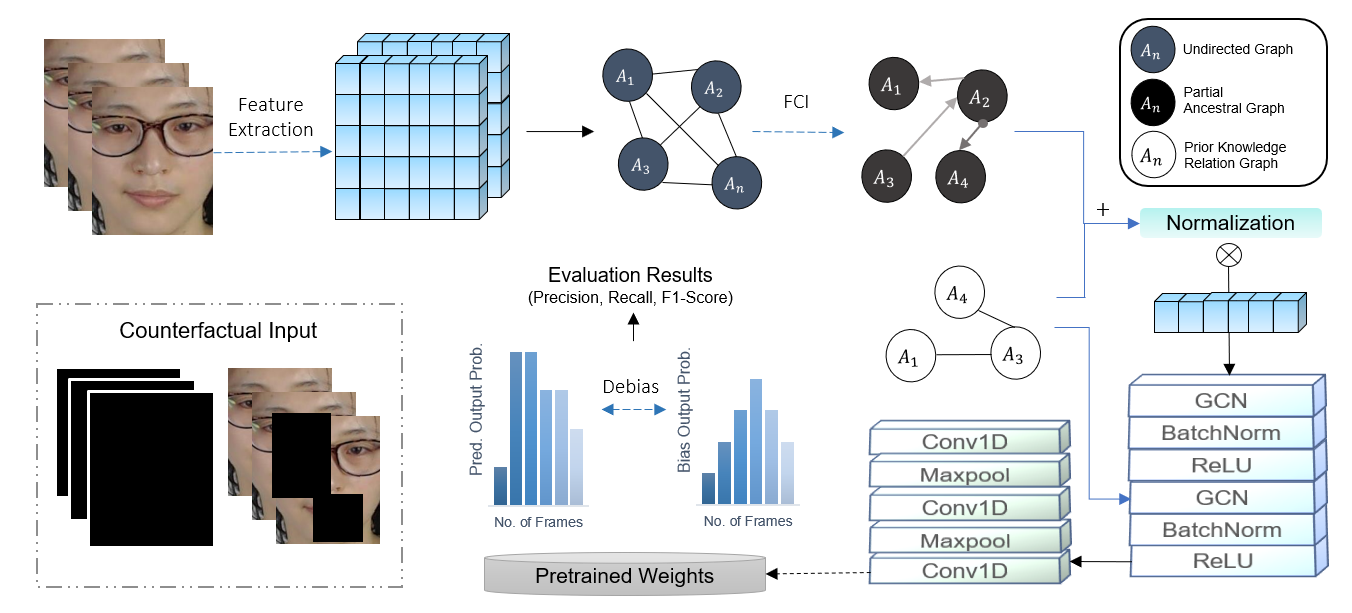}
    \caption{Illustration of all the proposed \textsf{Causal-Ex} and types of counterfactual inputs in this work.}
    \label{fig:arch}
\end{figure*}

\section{Methodology}
Our formal architecture consists of three modules, namely the data pre-processing stage, the causal modeling stage and the final debiasing stage. Figure~\ref{fig:arch} gives a brief overview of our method. The data pre-processing stage is largely inspired by~\cite{yin2023auaware} and it consists of cropping for face alignment facial landmark detection followed by extracting the optical flow and mean directional mean optical flow%feature
~(MDMO) features~\cite{liu2015main}. MDMO features track optical flow changes in essential Regions of Interest~(ROI) on the face, which are closely connected to facial action units, thus enhancing performance.  
We detail the rest of the modules %in detail 
below. 

\subsection{The Proposed \textsf{Causal-Ex}}
The data pre-processing stage outputs an adjacency graph of facial ROIs that are a combination of a set of AUs. This adjacency graph is an undirected graph  $D$ and purely captures the co-occurrence patterns. During stage two, the \textsf{Causal-Ex} module aims to learn the causal graph from the AU adjacency graph using the approach of~\cite{fci}.
\\

\begin{table*}[]\centering
\renewcommand{\arraystretch}{1.3}
\caption{Probability distribution of expression type and frames difference between two types of frames where $f_{onset}$, $f_{apex}$,$f_{offset}$ indicate the onset, apex and offset frames respectively. For instance, SAMM-LV contains 68.33\% micro expressions (ME) and 31.67\% macro expressions (MaE). On average, 31.7\% of videos have offset and onset frames within \textless101 frames (column 2), while in 59.2\% of videos, offset and apex frames are less than 51 frames apart (column 3). }
\begin{tabular}{c|clclccclccclccc}
\hline
\multirow{2}{*}{\textbf{Data}} & \multicolumn{3}{c}{\textbf{Exp.Type (\%)}} &  & \multicolumn{3}{c}{$\Delta$ (\textbf{$f_{offset}$,$f_{onset}$) (\%)}}                   &  & \multicolumn{3}{c}{$\Delta$ (\textbf{$f_{offset}$,$f_{apex}$) (\%)}}                       &  & \multicolumn{3}{c}{$\Delta$ (\textbf{$f_{apex}$,$f_{onset}$) (\%)}}                      \\ \cline{2-4} \cline{6-8} \cline{10-12} \cline{14-16} 
                                   & \textbf{ME}     &    & \textbf{MaE}  &  & \textbf{\textless{}31} & \textbf{31-60} & \textbf{\textgreater{}61} &  & \textbf{\textless{}21} & \textbf{21-40} & \textbf{\textgreater{}41} &  & \textbf{\textless{}11} & \textbf{11-20} & \textbf{\textgreater{}21} \\ \cline{1-2} \cline{4-4} \cline{6-8} \cline{10-12} \cline{14-16} 
\textbf{CAS(ME)$^2$}               & 18               &    & \textbf{82}   &  & \textbf{53.4}       & 35.5        & 11.1                   &  &  \textbf{57.1}       & 29.0        & 13.9                                      &  &  \textbf{48.2}       & 38.2        & 13.6                       \\
\multicolumn{1}{l|}{}              & \textbf{ME}     &    & \textbf{MaE}  &  & \textbf{\textless{}101} & \textbf{101-300} & \textbf{\textgreater{}300} &  & \textbf{\textless{}51} & \textbf{51-100} & \textbf{\textgreater{}100} &  & \textbf{\textless{}101} & \textbf{101-200} & \textbf{\textgreater{}200} \\ \cline{2-2} \cline{4-4} \cline{6-8} \cline{10-12} \cline{14-16} 
\textbf{SAMM-LV}                   & \textbf{68.33}   &    & 31.67         &  & 31.7       &\textbf{43.2}        & 25.1                  &  &  \textbf{59.2}	& 20.3 &	20.5

                                      &  &  \textbf{37.7} &31.7	& 30.6                      \\ \hline
\end{tabular} 
\label{dist}
\end{table*}

\noindent \textbf{Dynamic Action Units Graph.} 

Fast Causal Inference (FCI) is an algorithm that uncovers causal structures through a constraint-based approach. Given the adjacency graph $D$ representing face ROIs, FCI generates a partial ancestral graph (PAG) from observed video data. Unlike conventional causal graphs, PAGs incorporate both directed causal edges and undirected potential edges.
\label{sec:dynamicgrpah}
For a detailed mathematical formulation, proofs, and examples of the FCI algorithm, please consult \cite{fci}. PAG reveals four relationship types: (a) $V_m$ $\rightarrow$ $V_n$ indicates a causal link from $V_m$ to $V_n$ ($V_n$ being a child of $V_m$). (b) $V_m$ $\circ\rightarrow$ $V_n$ denotes the absence of an ancestral relation between $V_n$ and $V_m$, introducing unmeasured confounders $C$. These latent confounders impact both the independent (treatment) and dependent variable (outcome), potentially introducing bias and complicating the establishment of a causal relationship. (c) $V_m$ $\circ-\circ$ $V_n$ implies no $d$-separation between $V_n$ and $V_m$, signifying an unblocked path. (d) $V_m$ $\leftrightarrow$ $V_n$ suggests the presence of an unobserved confounder between $V_n$ and $V_m$. Specifically, we prefer relationships of type (a), showcasing direct causal relationships, represented by $V_m$ $\rightarrow$ $V_n$. This avoids involving indirect relations that might impact the model by introducing disturbance information.

The resulting PAG is a matrix $\text{A}_{12\times 12}$ containing the causal relationships between each of the 12 ROIs such that:
\begin{equation}
    {A}_{12\times 12}[i, j] = \begin{cases}
1, & \text{if } (i, j) \text{ is an edge in } PAG \\
0, & \text{otherwise}
\end{cases}
\end{equation}

In addition, as in \cite{yin2023auaware}, we also construct an AU-ROI relation matrix $B_{12\times 12}$. The computation of $B_{12\times 12}$ requires knowledge from the experts on the appropriate action units to be chosen and the consequent ROI-AU mapping known as the prior information. The matrix is formed through the collective knowledge of the entire training data such that:
   \begin{equation}
       {B}_{12\times 12}[i, j] =\sum_{\varphi k} \sum_{i,j} \mathbbm{1} (i \in f(U_i),\, j \in f(U_j))
   \end{equation}
where the $\varphi k$ is the $k$-th ground truth instances, $f(\cdot)$ acts as a mapping function, taking elements from a specific AU and associating them to the set of facial ROIs. The set $U_k$ represents the AUs present in the given video datasets. $U_i$ and $U_j$ are 
 subsets of $U_k$, the set of all AUs. Then, we obtain a compound matrix $C = A + B$ to be normalized and utilized in the subsequent steps.

\noindent \textbf{Spatio-Temporal Graph Embedding.} 
 This module learns a compact and meaningful representation capturing spatial-temporal patterns in the input features and the adjacency matric $C$ through convolutional  layers.

Let $H$ be the input feature matrix, $C$ the adjacency matrix, $(W^{(1)}, W^{(2)}, \ldots, W^{(L)})$ the learnable weight matrices for \(L\) GCN layers. The propagation equation for a multiple layer GCN is given by:
\begin{equation}
    H^{(l)} = \sigma\left({C} H^{(l-1)} W^{(l-1)}\right)
\end{equation}
where $ H^{(l)}$ is the node feature matrix at layer $l$, $H^{(l-1)}$ is the node feature matrix at layer $l-1$, $ W^{(l-1)}$ is the learnable weight matrix for layer $l-1$, $\sigma$ is the activation function, i.e. ReLU in our case because it helps capture the non-linear relationships in subtle facial expressions.

To capture the temporal features, we apply three layers of a $1-D$ convolutional network and the first two layers are added with a max-pooling layer to $H^{(l)}$. 

The output sequence $y$ from the convolutional layer at position $a$ is computed as:
\begin{equation}
    y_i = \sum_{b=1}^{k} (H_{ab} \cdot w_{ab})
\end{equation}
After the convolutional layer, the output of the max-pooling layer is denoted as $z$. Max-pooling takes a window of size $p$ and slides it across the input sequence to compute the maximum value within each window. For the $a$-th position in the output sequence from the convolutional layer, the output $z_a$ from the max-pooling layer is computed as:
\begin{equation}
    z_a = \max(y_a, y_{a+1}, \ldots, y_{a+p-1})
\end{equation}

The result is a downsampled sequence $z$ that captures the most salient features from the convolutional output. 

\subsection{Bias Removal Framework}
Publicly available long video datasets for macro and micro-expression spotting tasks currently exhibit limitations in terms of data volume and diversity. For instance, they often focus on mono-ethnic participants. We posit that this lack of diversity might contribute to bias in machine learning models.

\noindent \textbf{Dataset Bias.} In Table \ref{dist}, the last three columns show the distribution of frame differences between the three essential frames (onset, apex and offset) of the video i.e. how far apart they are, expressed by the number of frames between them. 
From Table \ref{dist}, it can be noted that the distributions of expressions in both datasets are highly imbalanced which may lead to the dataset bias. This is because over time, models learn an inductive bias by consistently predicting a given key frame i.e. any of the  $f_{onset}$, $f_{apex}$, $f_{offset}$ within a certain time interval or by associating a certain type of expression ME or MaE with spurious latent factors or dataset artifacts.
This indicates that the models rely on external factors unrelated to the inherent characteristics of the expressions for their classification. 

\noindent \textbf{Action Unit (AU) Bias.}
We also address AU bias in our proposed counterfactual debiasing method. The bias stemming from action units involves the probability of one or more action units being incorrectly correlated with a particular class. These erroneous correlations introduce biases wherein significant features, despite their importance, are overshadowed and play no role in determining the class. In micro-spotting, it is imperative to recognize and appreciate these critical differences.

\vspace{-0.3cm}
\subsection{Counterfactual Debiasing Method}
\begin{figure}[!ht]
    \centering
    \includegraphics[width=0.48\textwidth]{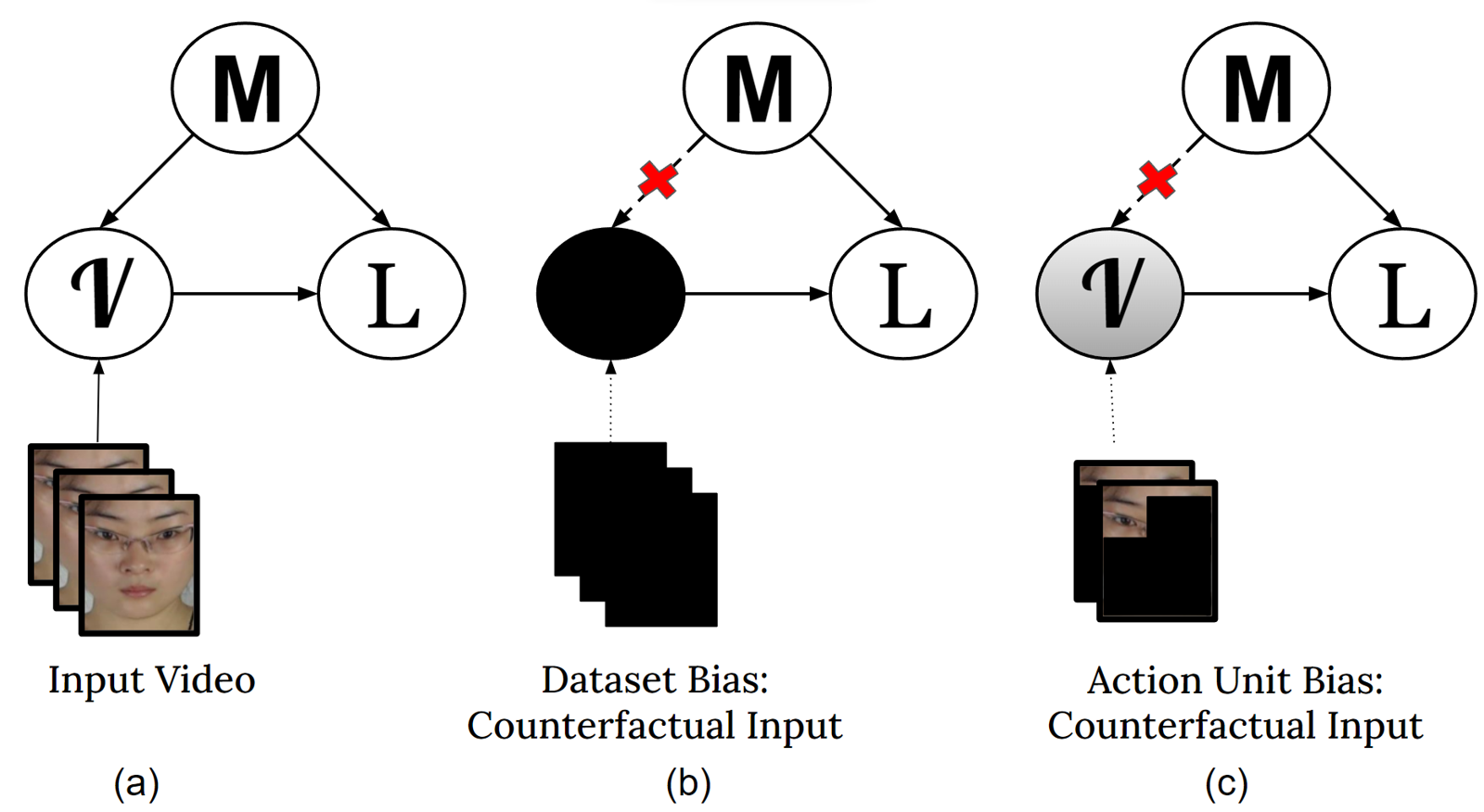}
    \caption{ \textit{\textbf{Counterfactual inference applied for video debiasing}}. (a) Video classification $M(l|\textbf{v})$ using a Pearl-like \cite{pearl2010causal} causal model that contains trained video classification model M, the input video $\textbf{v}\in$ V and corresponding ground truth label $l\in$ L. (b) Dataset bias is addressed by providing a counterfactual video with null frames as input to the model. (c) Action Unit bias is countered by the partially masked images with method in \cite{causally2024tan}}
    \label{fig:cause}
\end{figure}

In order to address the class-imbalance-based dataset bias, we extensively investigated by introducing controlled variations to the feature set $B$. Assume the causal structure %shown 
in  Figure~\ref{fig:cause} 
which shows a relation between the trained model $M$, the representation learned by the model $V$ and the expected outcome i.e. expression type $L$. Our objective is to expose the bias of model $M$ by inputting a non-informative video and observing its inherent biased prediction.

\begin{table*}[h]\centering
\renewcommand{\arraystretch}{1.1}
\caption{F1-score results of single database validation for baselines and \textsf{Causal-Ex}. Values in bold indicate the best results and the underlined are the second best among all results of the categories.}
\begin{tabular}{lccccclccc}
\hline
\multirow{2}{*}{Methods}  & &    & \multicolumn{3}{c}{CAS(ME)$^2$}                                                                           &  & \multicolumn{3}{c}{SAMM-LV}                                                                                  \\ \cline{4-6} \cline{8-10} 
                    & &          & Macro                               & Micro                      & \cellcolor{gray!15}Overall           &  & Macro                               & Micro                            & \cellcolor{gray!15}Overall           \\ \cline{1-5} \cline{5-10} 
MDMD \cite{mdmd}              & &           & 0.1196                              & 0.0082                     & \cellcolor{gray!15}0.0376           &  & 0.0629                              & 0.0364                           & \cellcolor{gray!15}0.0445           \\
STF \cite{stf}             & &             & 0.2131                              & 0.0547                     & \cellcolor{gray!15}0.1403           &  & 0.1331                              & 0.0725                           & \cellcolor{gray!15}0.0999           \\
Concat-CNN \cite{concatcnn1}      & &             & 0.2505                              & 0.0153                     & \cellcolor{gray!15}0.2019           &  & 0.3553                              & 0.1155                           & \cellcolor{gray!15}0.2736           \\
3D-CNN \cite{3D-CNNspot}             & &         & 0.0686                              & 0.0119                     & \cellcolor{gray!15}0.0497           &  & 0.1863                              & 0.0409                           & \cellcolor{gray!15}0.1193           \\
SOFTNet \cite{softnet}     & &                 & 0.2410                              & 0.1173                     & \cellcolor{gray!15}0.2022           &  & 0.2169                              & 0.1520                           & \cellcolor{gray!15}0.1881           \\
LSSNet  \cite{lssnet} & && \underline{0.3770}                        & 0.0420                     & \cellcolor{gray!15}0.325            &  & 0.2810                              & 0.1310                           & \cellcolor{gray!15}0.2380           \\
ABPN \cite{abpn}      & &                  & 0.3357                              & \textbf{0.1590}            & \cellcolor{gray!15}0.3117           &  & 0.3349                              & 0.1689                           & \cellcolor{gray!15}0.2908           \\
MTSN \cite{mtsn}        & &                 & 0.4104                              & 0.0808                     & \cellcolor{gray!15}\underline{0.3620} &  & 0.3459                              & 0.0878                           & \cellcolor{gray!15}0.2867           \\
Advanced Concat-CNN \cite{concatcnn}   & &                & 0.3179                              & \underline{0.1508}               & \cellcolor{gray!15}0.2839           &  & \underline{0.4049}                        & \textbf{0.2282}                  & \cellcolor{gray!15}\underline{0.3241} \\ \hline
\textbf{Ours (Before Debias)}& & & 0.4228                              & 0.1163                     & \cellcolor{gray!15}0.3852           &  & \multicolumn{1}{l}{0.4263}          & 0.1923                           & \cellcolor{gray!15}0.3693           \\
\textbf{Ours (After Debias)} & & & \multicolumn{1}{l}{\textbf{0.4241}} & \multicolumn{1}{l}{0.1364} & \multicolumn{1}{l}{\cellcolor{gray!15}\textbf{0.3880}} &  & \multicolumn{1}{l}{\textbf{0.4264}} & \multicolumn{1}{l}{\underline{0.2010}} & \multicolumn{1}{l}{\cellcolor{gray!15}\textbf{0.3701}} \\ \hline
\end{tabular}\label{comparative}
\vspace{-0.3cm}
\end{table*}

\noindent \textbf{Formation of Counterfactual Inputs.}
For dataset bias, specifically, we affected a transformation rendering these counterfactual features into null frames. \textit{Why do we choose this instead of a random values array or Gaussian blurs on the frames?} Random state often means ``pseudo-random" \cite{pseudorandom}, so it might still exhibit patterns or predictability in certain situations, while the Gaussian blurred images may still have some rough information \cite{gaussianresidual} on the subject's appearance and potentially lead to confusing counterfactuals. Thus, both of these do not fulfill the criteria of counterfactual inputs: i.e. not bringing any information to the model. 

Meanwhile, for the action unit bias, we apply the FCI algorithm to find the most influential ROIs in those particular testing datasets with the following equation:% below:
\begin{equation}
{E_f} = \{R_i \,|\, \forall R_j \in O(R_i) \geq O(R_j)\}
\end{equation}
In this equation, $E_f$ represents the set of essential ROIs, $R$ is the set of ROIs, $R_i$ is any region in $R$, $O(R_i)$ is the outdegree of $R_i$, and $O(R_j)$ is the outdegree of $R_j$. We masked all the facial regions except the $E_f$ as the counterfactual inputs to collect the bias vector of action unit bias. 

Both the dataset and action units counterfactual frames are denoted as $V_{0n}$, thus devoid of discriminative feature information. This manipulation is mathematically formalized as $M(L|do(V_n = V_{0n}))$, adhering to the conceptual framework previously articulated by~\cite{qian2021counterfactual}. An initial hypothesis posited that, in the absence of pertinent cues, the model, denoted as $M_\theta(.)$, would struggle to infer labels such as those for onset, apex and offset frames i.e. $f_{onset}$, $f_{apex}$ and $f_{offset}$ 
However, the experimental results unveiled a rather intriguing phenomenon, where the model continued to confidently predict these labels even for instances marked by the counterfactual frames. This compelling observation suggests an implicit encoding of biases within the model's learned representations.

In our methodology, both $M(L|V_n)$ and $M(L|do(V_n = V_{0n}))$ yield probability distributions $P(.)$ for each frame, encapsulating the likelihood of classification into the different frame categories within $L$. Formally, denoted as:
\begin{equation}
P_d = {[ h_{on, n}, h_{off, n}, h_{apex, n}, h_{type, n} ]}
\end{equation}
where $d$ denotes the testing input type, $n$ is the number of frames, and $h_{on, n}, h_{off, n}, h_{apex, n}, h_{type, n}$ represent probabilities corresponding to the onset, offset, apex, and expression type classifications, respectively

In response to these observations, we propose a novel approach to bias mitigation leveraging Pearl's back-door criterion~\cite{causalstat}. Concretely, we begin by recording the original biased probabilities, denoted as $P_{V_n}$, for a given input video $V_n$. Subsequently, we effectuate a debiasing mechanism by subtracting $P_{V_{0n}}$ from $M(L|V_n)$, yielding an final adjusted probability distribution as $F$:
\begin{equation}
\begin{split}
& F(l|\textbf{v}) = M(L|V_n) - \lambda_{param} M(\hat{L}|do(V_n=\hat{V_{0n})})
\end{split}\label{debias}
\end{equation}
In the above equation, $\lambda_{param}$ serves as a hyperparameter, modulating the magnitude of debiasing. Notably, distinct $\lambda_{param}$ values are tailored to each $P$ set, ensuring the relevance and effectiveness of the debiasing procedure. This counterfactual debiasing methodology is conducted during the inference phase, offering a practical, interpretable, and computationally feasible strategy for addressing biases that may distort the model's decision-making process.
\vspace{-0.5cm}
\section{Experiments}
\noindent
\textbf{Datasets.} Our proposed method utilizes the following long-video datasets for training and testing, commonly used in existing spotting models. \textbf{CAS(ME)$^2$} \cite{casmesquare} comprises 303 video samples, including 250 spontaneous macro-expressions and 53 micro-expressions. Emotion labels are based on Action Units (AUs), self-reported facial movements, and stimuli used to evoke specific emotions. \textbf{SAMM-Long Video} \cite{sammlv} contains 147 long video clips with 159 micro-expressions and 343 macro-expressions, annotated using the Facial Action Coding System (FACS). All clips have a resolution of 2040 × 1088. 

Emotion labeling criteria stem from FACS coding, observed emotion categories, and self-reported emotions. However, we cannot rule out potential bias and lack of causality awareness in the data collection process for both datasets.

\noindent \textbf{Implementation Details.} The number of training epochs in the experiments is 100 with the Adam optimizer. After analyzing different sets of epoch numbers, 100 epochs is sufficient because both datasets are relatively small in sample size, the input graph is not too large, and the network architecture and hyperparameters are well-tuned.

\emph{Leave-One-Subject-Out (LOSO)} is adopted in our validation process to ensure no overfitting occurs. We measured and validated our results with \emph{F1-score}, which measures the accuracy and balance between precision and recall of a classification model. This metric is commonly used in imbalance-label datasets. Imbalance is a typical characteristic of facial expression datasets as the emotional expressions are naturally induced, unfortunately, some emotions are harder to be naturally expressed or triggered. Implementations utilize Python 3.9.13, PyTorch 2.0.1, and CUDA 11.7 with Nvidia A-100 GPUs on a SLURM system.\subsection{Model Performance} 

\begin{table}[ht]\centering
\renewcommand{\arraystretch}{1.1}
\caption{Results on GCN-based models vs \textsf{Causal-Ex} with and without our proposed dynamic causal relation graph embedding module and debiasing module. (\cmark: with , \xmark : without )}
\begin{tabular}{ccccc}
\hline
\multicolumn{5}{c}{CAS(ME)$^2$}                         \\ \hline \hline
Methods                    & Prior & Causal & Debias & F1-Score        \\ \hline
\multirow{2}{*}{AUW-GCN}   & \cmark     & \xmark      & \xmark      & 0.3739          \\
                           & \cmark     & \xmark      & \cmark      & \textbf{0.3774} \\ \hline
\multirow{2}{*}{\textsf{Causal-Ex}} & \cmark     & \cmark      & \xmark      & 0.3852         \\
                           & \cmark     & \cmark      & \cmark      & \textbf{0.3880} \\ \hline
\multicolumn{5}{c}{SAMM-LV}                                            \\ \hline \hline
Methods                    & Prior & Causal & Debias & F1-Score        \\ \hline
\multirow{2}{*}{AUW-GCN}   & \cmark     & \xmark      & \xmark      & 0.3767          \\
                           & \cmark     & \xmark      & \cmark      & \textbf{0.3781} \\ \hline
\multirow{2}{*}{\textsf{Causal-Ex}} & \cmark     & \cmark      & \xmark      & 0.3693          \\
                           & \cmark     & \cmark      & \cmark      & \textbf{0.3701} \\ \hline
\end{tabular}\label{gcncompare}
\vspace{-0.4cm}
\end{table}

\noindent
\textbf{Single-Database Validation.} We compared our proposed methods with a few notable baselines discussed in the related work. Evaluation results for CNN-based models and hand-crafted methods are concluded in Table \ref{comparative}. Table \ref{gcncompare} shows the comparison results of AUW-GCN \cite{yin2023auaware}~(the other only known AU-based spotting method) and our models w/ and w/o debiasing. Some of the baselines' codes are publicly available and we included the reproduced results for these baselines. Our method outperforms the SOTA baselines in the comparative studies of Table \ref{comparative}. The F1-score %rises 
increases by 2.6\% and 4.6\% in overall performance for CAS(ME)$^2$ and SAMM-LV, respectively. While for the GCN-based model, we applied the debiasing module on the AUW-GCN model and an increase in F1 score is observed. After adding the causal graph to the GCN model, CAS(ME)$^2$ performs better than the existing work, increasing 1.06\% in the metrics score. 

\noindent
\textbf{Cross-Database Validation.} Besides the single dataset that previous literature worked on, we also evaluate our model with cross-dataset validation as shown in Table \ref{com_cross} with and without prior knowledge graph. Testing the dataset samples with the pre-trained models of different datasets could show the effectiveness and robustness of our methods. The CAS(ME)$^2$ dataset has a larger sample size compared to SAMM-LV. This may be the reason that the pre-trained CAS(ME)$^2$ model has a notable performance by testing with SAMM-LV. When we have both causal graph and prior knowledge graph in the cross-validation testing, the F1-score results of the micro-expression category for SAMM-LV have a higher value when testing with the CAS(ME)$^2$
pre-trained weights.

\begin{figure}
    \centering
    \includegraphics[width=.4\textwidth]{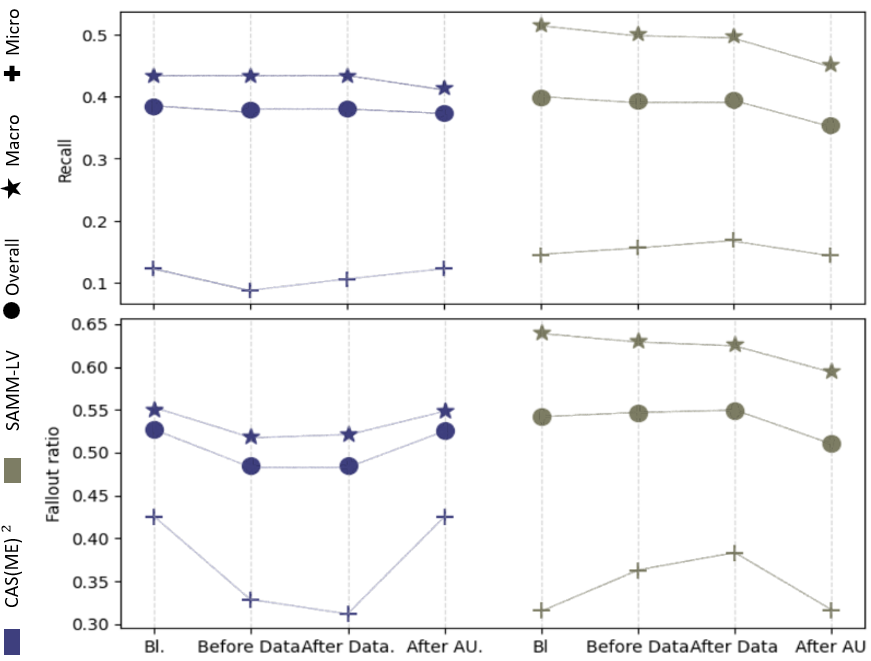}
    \caption{ A comparison of Recall~(TPR) on the \textbf{top} vs Fallout Ratio~(FPR) on the \textbf{bottom} for micro and macro expression classes in  CAS(ME)$^2$ and SAMM-LV datasets. The four lines indicate the estimate on Baseline model, Before debiasing, after dataset debiasing, and finally  Causal-Ex after action unit debiasing. Ideally, we prefer a balanced $EO = \frac{TPR}{FPR}$ for each of the classes.}
    \label{onlycausal}
\end{figure}

\begin{figure}
    \centering
    \includegraphics[width=.5\textwidth]{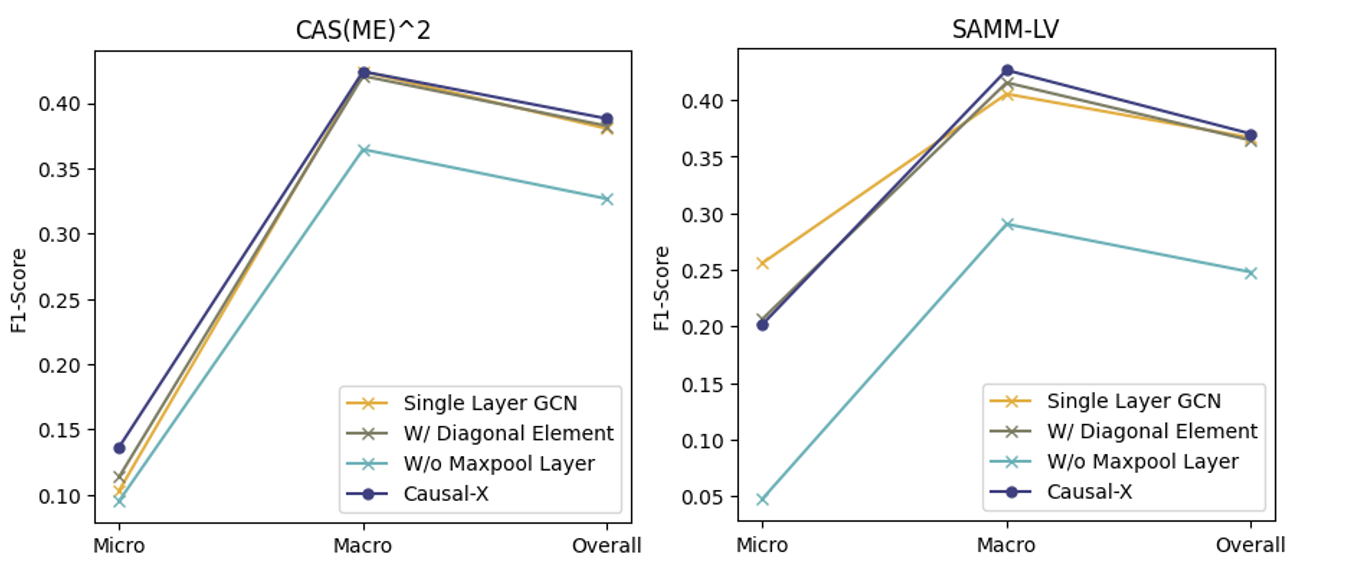}
    \caption{Ablation studies on both datasets.}
    \label{ablative}
    \vspace{-0.5cm}
\end{figure}

\begin{table*}[]\centering
\renewcommand{\arraystretch}{1.1}
\caption{Cross database generalization results on \textsf{Causal-Ex}. $Pt.$ stands for the pretrained model and $Test$ is the testing datasets.}
\begin{tabular}{clclclclclclc}
\hline \multirow{2}{*}{\diagbox[width=7em]{Pt.}{Test}}
  &  & \multicolumn{5}{c}{CAS(ME)$^2$} &  & \multicolumn{5}{c}{SAMM-LV}     \\ \cline{3-7} \noalign{\vskip\doublerulesep
         \vskip-\arrayrulewidth} \cline{3-7} \cline{9-13} \noalign{\vskip\doublerulesep
         \vskip-\arrayrulewidth} \cline{9-13}
&  & Macro-Exp                       &  & Micro-Exp                       &  & Overall                         &  & Macro-Exp                       &  & Micro-Exp                       &  & Overall                         \\ \cline{1-3} \cline{5-5} \cline{7-7} \cline{9-9} \cline{11-11} \cline{13-13} 
CAS(ME)$^2$ (Causal)                                                             &  & \textbf{0.4238}   &  & \textbf{0.0723}      &  & \textbf{0.3792}                       &  & 0.3554                          &  & 0.0220                           &  & 0.2797                           \\ \cline{1-2}
SAMM-LV (Causal)                                                                 &  & 0.1942                         &  & 0.0233 &  & 0.1780 &  & \textbf{0.4145}   &  & \textbf{0.2035}      &  & \textbf{0.3692} \\ \hline
CAS(ME)$^2$ (Causal+Prior)                                                             &  & \textbf{0.4241}   &  & \textbf{0.1364}      &  & \textbf{0.3880}                       &  & 0.3874                           &  & \textbf{0.2143}                           &  & 0.3560                           \\ \cline{1-2}
SAMM-LV  (Causal+Prior)                                                                  &  & 0.1950                         &  & 0.1190&  & 0.1878 &  & \textbf{0.4264}      &  & 0.2015     &  & \textbf{0.3701}\\ \hline
\end{tabular}\label{com_cross}
\vspace{-0.6cm}
\end{table*}

\subsection{Ablative Study}
\noindent
\textbf{Impact of Causal Relation Graph.} To our knowledge, our model is unique in integrating causal knowledge and addressing bias effects in expression spotting. We conduct the experiments on both datasets by ignoring the ground-truth expert annotated AU-ROI relations. The causal graph is solely responsible for building the adjacency matrix in the GCN. Solely leveraging the causal structure in the input graph for GCN yields superior results compared to many baselines. Causal graphs empower GCN for counterfactual reasoning, simulating variable interactions. With causal knowledge, GCN captures significant patterns and dependencies, enhancing predictive accuracy and model interpretation. To verify the robustness of our method in handling bias, we explore the criterion of Equalized Odds (EO)\cite{equalizeodd}.
A model is considered fair when its predictions exhibit uniformity in both True Positive Rate (TPR) and False Positive Rate (FPR) across all groups within the dataset. EO in our context can be written as: 
\vspace{-0.3cm}
\begin{flalign}
    P(\hat{M} = 1 | M = m, I = g_0) &= 
    P(\hat{M} = 1 | M = m, I = g_n) &
\end{flalign}

where is $\hat{M}$ is the predicted value, $m \in \{0,1\}$ is the ground truth, and $I \in [g_0, g_n]$ represents the group membership. In this task, the groups are baseline, \textsf{Causal-Ex} without debiasing and \textsf{Causal-Ex} after dataset and action unit debiased. In Figure \ref{onlycausal}, our TPR and FPR for different groups are comparable, aligning with the criterion of EO while accounting for practical considerations and the inherent trade-offs involved in achieving fairness.

\begin{figure}[h]
    \centering
    \includegraphics[width=.45\textwidth]{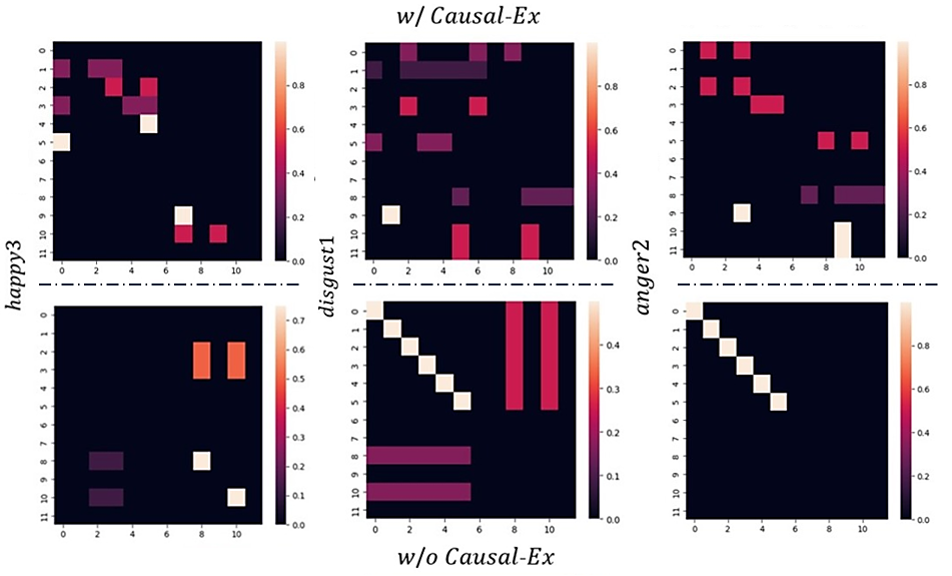}
    \caption{FCI algorithm-generated Relation Graph for subject s16's 12 ROIs in CAS(ME)$^2$ (left) vs a graph from prior knowledge (right) reveals that the latter contains multiple null values, enhancing feature refinement but loses much input information.}
    \label{fig:casualbyemo}
    \vspace{-0.3cm}
\end{figure}

\noindent
\textbf{Illustration of Correlation between ROIs Nodes.} Figure \ref{fig:casualbyemo} illustrates a clear distinction between causal relationships and ground truth labels in the correlation graph of ROI nodes. The weights of the nodes and edges convey critical relational information, with higher color intensity in the heatmap indicating greater weight. \textbf{Case Studies.} To identify the optimal model for the causal relation graph, we conducted experiments with three ablation reviews, as shown in Figure \ref{ablative}. The \textsf{Causal-Ex} model with two GCN layers captures complex patterns in long videos that a single layer cannot fully represent. Additionally, the \textsf{Causal-Ex} model with an adjacency matrix excluding self-loop functions performs better when omitting diagonal elements, i.e., cases where a node influences itself. Finally, the \textsf{Causal-Ex} model without the max-pooling layer shows that incorporating a max-pooling layer after each convolutional layer enhances the model's ability to spot micro-expressions.

\noindent
Limitations persist in this study, as causal graph construction depends on input data and its size, yielding mixed direct and indirect edges in the PAG relation, and unmeasured confounders might influence relationships. Future research could involve incorporating counterfactual studies to understand the AU-ROI relationships better. 
\vspace{-0.5cm}
\section{Conclusion}

In this work, a novel method is proposed to generate a causal relation graph between the ROIs to bring more information on edge and nodes for the features. The spatiotemporal graph embedding module enables %helped 
the model to learn a better representation of every frame with the dynamic causal graph. Furthermore, %To 
we take Pearl's fairness approach \cite{pearl2010foundations} and designed a counterfactual debiasing module in our model to ensure and supervise the fairness. In this proposed work, we emphasise the causality and bias in the task and tackle them in in-propcessing and post-processing. The outcomes outperform the state-of-the-art for both CAS(ME)$^2$ and SAMM-LV datasets. 

% \printcredits
\vspace{-0.5cm}
\bibliographystyle{model1-num-names}
\bibliography{refs}

\end{document}